\documentclass{tMPH2e}   

 \usepackage{graphicx}
 \DeclareGraphicsExtensions{.eps}
\usepackage{subfigure}
\usepackage{amsmath}
\usepackage{algorithm}
\usepackage{algorithmic}
\usepackage{paralist}
\begin{document}
\title{ Group Leaders Optimization Algorithm}
\author{Anmer Daskin$^{a}$ and Sabre Kais$^{b}$ $^{\ast}$\thanks{$^\ast$Corresponding author. Email: kais@purdue.edu
\vspace{6pt}}\\\vspace{6pt}  $^{a}${\em{Department of Computer Science, Purdue University, West Lafayaette, IN, 47907, USA}};
$^{b}${\em{Department of Chemistry and Birck Nanotechnology Center, Purdue University, West Lafayette, IN 47907, USA}}}

\maketitle
\begin{abstract}

We present a new global optimization algorithm in which the influence of the leaders
 in social groups is used as an inspiration for the evolutionary technique which is designed into 
a group architecture. 
To demonstrate the efficiency of the method, a standard suite of  single and
 multi dimensional optimization functions along with
the energies and the geometric structures of Lennard-Jones clusters  are given as well 
as the application of the algorithm on quantum circuit design problems. 
We show that as an improvement over previous methods, the  algorithm scales as $N^{2.5}$ for the Lennard-Jones clusters of N-particles. 
In addition, an efficient circuit design is shown for two qubit Grover search algorithm 
which is a quantum algorithm providing quadratic speedup over the classical counterpart.  
\end{abstract}

\section{Introduction} 
Global optimization is one of the most important computational problems in science and engineering.
Because of the complexity of optimization problems and the high dimension of the search space, 
in most cases, using linear or 
deterministic methods to solve them may not be a feasible way \cite{W2009GOEB}. 
Wille and Vennik\cite{Wille} argued that global optimization  of a cluster of identical atoms interacting under 
two-body central forces, belong to the class of NP-hard problems. This means that as yet 
no polynomial time algorithm solving this problem is known. Recently, Adib \cite{Adib} 
reexamined the computational complexity of the cluster optimization problem and suggest that the 
original NP level of complexity does not apply to pairwise potentials of physical interest, such as those that depend on 
the geometric distance between the particles. A geometric analogue of the original problem 
is formulated and  new subproblems that bear more direct consequences to the
 numerical study of cluster optimization where suggested. However, the intractability of this 
subproblem remains unknown, suggests the need for good heuristics.

Many optimization methods have been developed and then can be largely classified into
two groups, deterministic and stochastic. Deterministic methods include variations on Newton's
method such as discrete Newton, quasi Newton  and truncated Newton\cite{Dennis,Gilbert}, 
tunneling method\cite{Levy}
and renormalization group methods\cite{Shalloway}. Stochastic methods include, simulated annealing\cite{Kirkpatrick},
quantum annealing\cite{Finnila}, J-walking\cite{Frantz}, tabu search\cite{Cvijovicacute}, genetic algorithms\cite{Goldberg} 
and basin-hoping approach \cite{Wales27081999, wales-1997-101}.
More recent work on  probabilistic techniques have been proposed to solve these optimization 
problems by observing  nature and modeling social behaviors and characteristics. Including
genetic algorithms (GA), evolutionary algorithms (EA) such as the particle swarm optimization 
algorithm (PSO) and the Pivot methods\cite{W2009GOEB,229867,PhysRevE.55.1162,Kennedy95,
StantonBleilKais,SerraStantonKais,Nigra1999433}.

Implementation of many of these algorithms on complex problems requires exhausting 
computational time 
and a growing need for more computer resources depending upon: 
the dimension, the solution space and the type of the problem.
 The key to speed up the optimization process is reducing the number of computations
 in the algorithms while keeping the amount of iterations low and the success 
rate of the algorithms high.
This paper introduces a new global optimization algorithm which reduces the 
optimization time, and is both simple and easy to implement.
  In the following sections, the inspiration and the implementation  of the algorithm 
will be explained and
  test results will be given  for  some of the most famous optimization test problems; for the 
global optimization of the minimum energy structures
 of complex  Lennard Jones clusters; and for the quantum circuit design of the Grover search algorithm.

\section{Inspirations and Related Works}
Leaders in social groups affect other members of their groups by influencing either the number of 
members or each member intensively \cite{Weinberg}. 
Therefore, the effect of group leaders inclines the groups to have uniform behavior and  
characteristics similar to the leader.
 These new behaviors and characteristics may improve or worsen the quality of the members of a 
group. A leader represents the properties of its group. 
To become a leader for a group requires a person to have some better abilities than  others 
in the group.  

Similar ideas to using leaders or/and grouping solution population have been the inspiration for 
optimization algorithms such as Cooperative Co-evolutionary Genetic Algorithms (CCGA), 
Cooperative Co-evolutionary Algorithms(CCEAs) \cite{Potter94acooperative,citeulike:3182909}, 
and Parallel Evolutionary Algorithms(PEAs) \cite{517269}.  
However,  instead of trying to fully simulate the influence of 
leaders on their peers in social groups by 
constructing a population which includes small, interacting groups with their leaders, 
most of these and other similar algorithms have attempted the decomposition of big and complex problems 
into subcomponents or 
divide the whole population into multiple subpopulations with a parallel structure. 
In CCGAs, as described  by Potter et al. \cite{Potter94acooperative}, each species -which 
are evolved into subcomponents 
by using a standard genetic algorithm- represents a subcomponent of the potential solution, and
 each representative member of the species is used to form the complete solution of the optimization problem. 
In \cite{1108890}, a general architecture for the evolving co-adapted subcomponents was presented
 in order to  apply evolutionary algorithms to complex problems. Therefore, instead of GA, 
Particle Swarm Optimization Algorithm has been used for the subcomponents of Potter's CCGA structure by van den Bergh et al. \cite{citeulike:3182909}.
Briefly, the general framework of cooperative coevolutionary algorithms has three main steps: problem decomposition, subcomponent optimization, 
and subcomponent adaptation \cite{Yang20082985}.  

In PEAs the whole population forms in a distributed way and consists multiple subpopulation.
Single-population master-slaves, multiple populations, fine-grained and hierarchical combinations 
are the main types of PEAs \cite{WinGeiGal}. The proposed algorithm in this paper differs from
the PEAs  in that all members of the population are interacting and 
there is a mutual effect between the members of the population in addition to leaders' effect on 
individuals in their groups. The sum of all these interactions 
forms the evolutionary technique. However,
in PEAs, in most cases, the interaction between subpopulations is made with the migration of 
individuals and evolutionary techniques 
used for subpopulations can be independent from each other. 
 
\section{Group Leaders Optimization Algorithm}
\subsection{General Idea}

Inspired by leaders in social groups and cooperative coevolutionary algorithms, 
we have designed a new global optimization algorithm in which there are separate groups and leaders for each group. Initially forming groups does not require members to have some similar characteristics. Instead, it is based on random selection. 
While CCGA and other similar algorithms decompose the solution space, and each group represents a solution for a part of the problem, 
in our algorithm each group tries to find a global solution by being under the influence of the group 
leaders which are the closest members of the groups to local or global minima. 
The leaders are those whose fitness values are the best in their groups,
 and a leader can lose its position after an iteration if another member in the same group becomes having a better fitness value. 

Since in social networks, leaders have effects on their peers, thusly the algorithm uses  the some portion of leaders while generating new group members. 
Hence, a leader, (in most cases a local optimum) dominates all other solution candidates (group members) surrounding it, 
and the members of a group come closer and resemble their leader more in each iteration.  
By this way, the algorithm is able to search the solution space between a leader and its group members thoroughly, 
and so is able to search the area for a local or a global optimum (or an approximation of it) in a fast way.

After a certain number of evolutions, it is obvious that the members may become too similar to their leaders. 
To maintain the diversity of the group, for each group, we transfer some variables from different groups by choosing them randomly. 
In addition to providing diversity, this one way crossover helps a group to jump out of local minima and search new solution spaces. 

\subsection{Algorithm Steps}
\begin{figure*}[!t]
\centering
\includegraphics[width=0.9\textwidth]{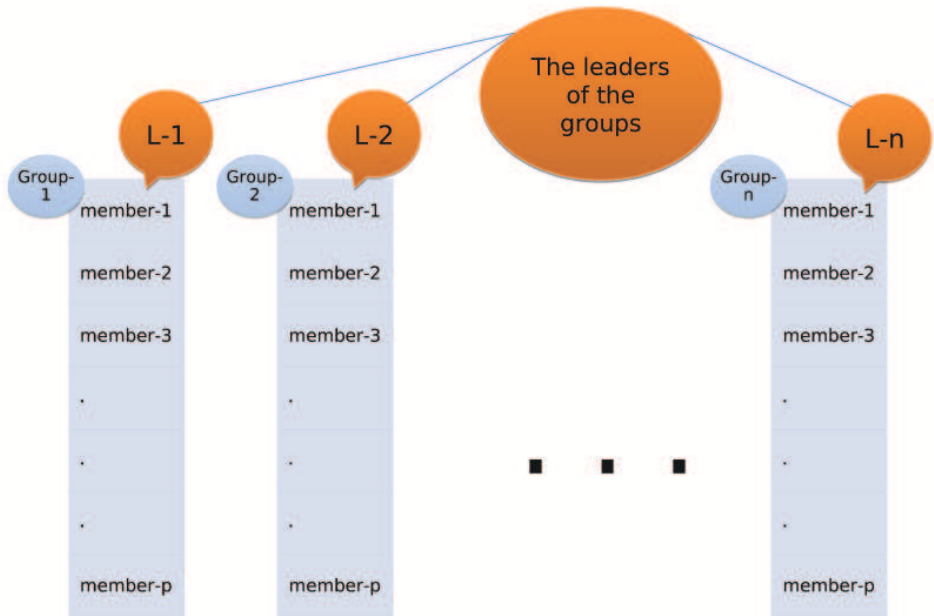}
\caption{Steps 1-3 of the algorithm; groups consisting of p number of members are created, and their leaders are chosen based on the fitness values.} 
\label{step1-3}
\end{figure*}

\begin{figure*}[!t]
\centering
\includegraphics[width=0.9\textwidth]{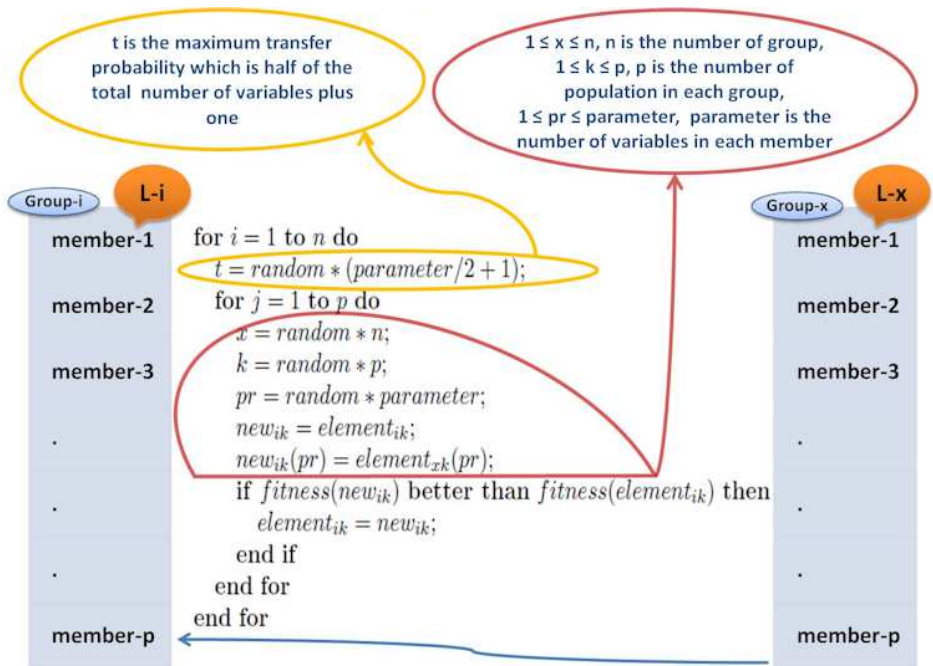}
\caption{ Step 5 of the algorithm: one way crossover:  
$pr$th variable of an element, member of the $i$th group, is replaced by $pr$th variable of the $k$th member of $x$th group. 
The same operation is repeated t times for each group (maximum -half of the number of variables plus one- times for each group). 
The arrow at the bottom of the figure shows the direction of transfer. }
\label{step4-6}
\end{figure*}

In this section,  algorithm steps are described with their reasons in sequence and are shown  in Fig.\ref{step1-3} and Fig.\ref{step4-6}.

\textit{ \textbf{Step 1: Generate p number of population for each group randomly:}} The total population for each group is p, hence, the whole population is $n*p$ where n is the number of groups. Creation of the groups and the members are totally random.

\textit{\textbf{Step 2: Calculate fitness values for all members in all groups:}} All solution candidates, group members, are evaluated in the optimization problem and their fitness values are assigned. 

\textit{\textbf{Step 3: Determine the leaders for each group:}} Each group has one leader and the leaders are ones whose fitness values are the best 
within their respective groups. The algorithm steps 1-3 are shown in the Fig. 1.

\textit{\textbf{Step 4: Mutation and recombination:}} Create new member by using the old one, 
its group leader, and a random element. If the new member has better fitness value than old one, 
then replace the old one with the new one. Otherwise, keep the old member.
For numerical problems, the expression simply reads; 
\begin{equation}
\label{e_rate}
new = r1*old + r2*leader + r3*random.
\end{equation}
In Eq.(\ref{e_rate}), $r_1$, $r_2$, and $r_3$ are the rates determining the portions of old (current) member, 
leader, and random while generating the new population. Although in this paper, $r_1$, $r_2$, and $r_3$ will always sum to 1, it is not a requirement for the algorithm in general.    
For instance, let the current element be 0.9, the leader  1.8, and the generated random element 1.5, and suppose  $r_1=0.8$, $r_2=0.19$, and $r_3=0.01$. In this case, the new element is equal to 1.077. Then fitness values of the old and the new element are checked.
If fitness(1.077) is better than the fitness(0.9), then the old element  is replaced by the new one. Pseudo-code for this step is as follows:
\begin{center}
\begin{flushleft}
\begin{algorithmic}
\FOR {$i = 1$ \TO $n$}
\FOR {$j=1$ \TO $p$}
\STATE $new_{ij}=r_{1} * member_{ij} + r_{2} * L_{i} + r_{3} * random;$
\IF {$fitness(new_{ij})$  \textbf{better than} $fitness(member_{ij})$} 
\STATE $member_{ij}=new_{ij};$
\ENDIF
\ENDFOR
\ENDFOR
\end{algorithmic}
\end{flushleft}
\end{center}
 In the pseudo-code:
 n is the number of groups, p is the number of population in each group, $r_1, r_2$ 
and $r_3 $ are the rates of old value of the members of groups, leaders,
 and the random part.The expression in the inner loop is the general formula
 of recombination and mutation for numerical optimization problems.

Depending on the value of $r_1$ an element stays its original characteristics, and depending on the value of $r_2$ it becomes more alike 
its leader during iterations. Thus, in some cases, choosing the right values for  $r_1, r_2$ and $r_3 $ may play an
 important role for the algorithm to get better results during the optimization. However, choosing these parameters by obeying the property,
 $r_3 , r_2 \le 0.5 \le r_1$ allows one to have a thorough search of a solution space. Hence, this minimizes the effect of these parameters on the results.
    
The main benefit of this evolution is that the algorithm becomes able to search the solution space surrounding the leaders (which are possibly local or global minima). 
Therefore, this allows the population to converge upon global minima in a very fast way. 
The members of the groups are not subjected to a local minimization; however, an implementation of Lamarckian concepts of evolution 
for the local minimization \cite{Turner2000183} gives a more stable and efficient algorithm. 

It is also important to note that Eq. (\ref{e_rate}) looks similar to the updating equation of PSO \cite{Kennedy95}. 
The difference is that a member is always at its best position and the best position of a member is not saved in a parameter as is done in PSO, 
hence there is no information about the member's(or the particle's) previous positions (values). 

\textit{\textbf{Step 5: Parameter transfer from other groups (One way crossover):}}  Choose random members starting from the first group,
 and then transfer some parameters by choosing another random member from another group. If this transfer makes a member have a better fitness value, then change the member,
 otherwise keep the original form. This process is shown in Fig.\ref{step4-6} via pseudo code. This one-way crossover has similarities with the difference vector of Differential Evolution\cite{DifEvolution}.
 The difference is that the transfer operation is between the members which are in different groups.
In this step, it is important to determine correct transfer rate, otherwise all populations may quickly become similar.
 In our experiments, transfer operation rate was taken  t times (t is a random number between 1 and half of the number of total parameters (variables) plus one ($1\le t \le \frac{\sharp parameter}{2}+1$)) for each group (not for each member). And each time, only one parameter is transferred.

\textit{\textbf{Step 6: Repeat step 3-step 5 number of given iteration times}}.

Since each group looks for the solution in mostly different spaces, 
GLOA is able to search different solution spaces simultaneously.
 We did not put any constraint for groups to only search in subspaces, so a few groups may also search the same places. 
However, this does not make them redundant as they allow GLOA to find different local or global minima within the same subspace.
 Since each group has a leader and the leaders direct the other members of the group in order to search the area between 
the leader and the members of the group, it is able to search for a good solution (around of the leader of the group). 

In addition to increasing the level of diversity of the groups, transferring some parameters (crossover) between groups allows the algorithm 
to direct the members of a group to search different spaces. Therefore, if a group has found some parameters correctly or very close to correct, 
then transferring parameters between groups allows other groups to get these parameters and find their solutions faster.  Since it is transferred only 
parameters which make the member have better fitness value, the spreading of a member who has a bad fitness value is avoided. 
In terms of optimization problems requiring highly diverse populations,  choosing to do crossover and mutation-recombination steps without comparing fitness 
values may be wise and may improve the effectiveness of the algorithm. 

\section{Optimization  Results}

The parameters for the algorithm have effects on the quality of results. As it is stated in the previous section, 
choosing $r_3 , r_2 $ less than and $r_1$ greater than 0.5
 makes the algorithm more stable and minimizes the effects of these parameters. 
The number of groups and the population of the groups should be chosen to be large enough depending on the complexity of the problem. 
Therefore, while generating new elements, the rate of crossover between groups and the portions of elements, leaders,
 and random elements should be carefully adjusted for the type and the complexity of the optimization problem.
 For instance, if one takes crossover rate or the portion of the leaders too high in relation to the chosen group and population number,
 this may cause the whole population to become uniform very quickly. Hence, the algorithm may get stuck in a local solution, and not search the whole solution space. 
The algorithm was tested on different type of optimization problems, one dimensional 
and multi dimensional optimization test functions, and it was also used to
 find the minimum energy structures of Lennard Jones Clusters. 

\subsection{Test Functions}

While testing optimization algorithms on numerical problems, search domain and the number of iteration have crucial effects on the performance of the algorithms.
In terms of the implementation, the number of groups and the population were taken the same for all numerical problems where the number of groups is 10,
 and the population number in each group is 25. Keeping the number of leaders the same for all numerical test functions which have different numbers of local minima 
shows the capability of the algorithm to escape from local minima is not highly related to the number of leaders.
 While  results are shown in terms of function value and the number of iteration in graphs for single dimensional 
test problems, for multi dimensional problems, they are presented in terms of the number of dimension (up to 1000) and the minimum function value in each dimension.

\subsubsection{Single Dimensional Test Problems}
For the single dimensional test functions, the parameters of the algorithm are shown in Table-\ref{tsingle}. 
In this  table, $r_3=0$,  that means that there is no randomness while generating new populations. 
We observed from our optimization  that this makes GLOA converge to a global minima faster for single dimensional test problems. 

\begin{figure*}[!t]
\centerline
\hfil
\subfigure[Beale Function]{\includegraphics[width=2.5in]{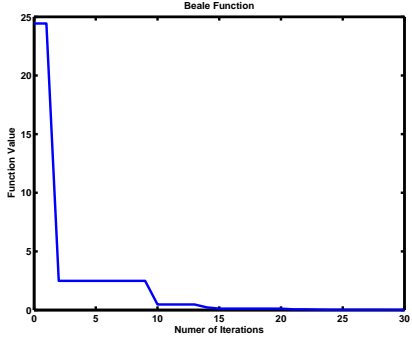}
\label{fig_beale}}
\hfil
\subfigure[Easom Function]{\includegraphics[width=2.5in]{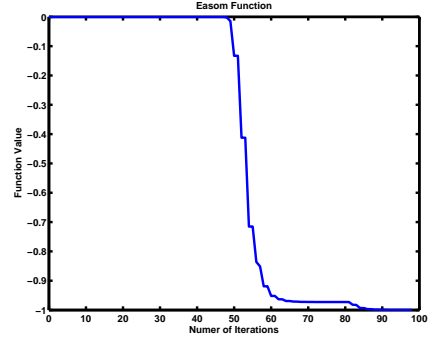}
\label{fig_easom}}

\subfigure[Goldenstein-Price's Function]{\includegraphics[width=2.5in]{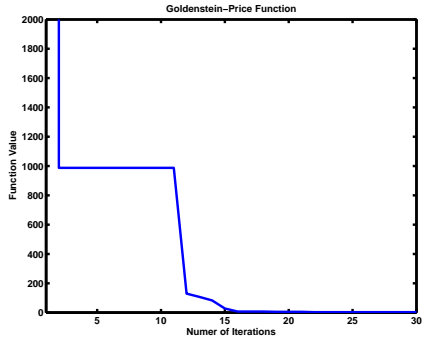}
\label{fig_gold}}
\hfil
\subfigure[Shubert Function]{\includegraphics[width=2.5in]{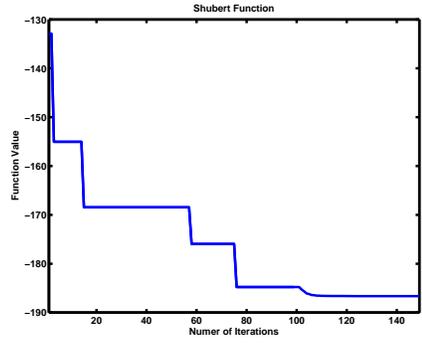}
\label{fig_shubert}}
\caption{Results For Single Dimensional Optimization Test Functions: x-axis represents the number of 
iterations and y-axis is the function value which is the result of the best member in the population. 
a)The Result for Beale Function b)The Result for Easom Function c)The Result for Goldein-Price's Function d)The Result for Shubert Function}
\label{singletest}
\end{figure*}

Also, as shown in Fig.\ref{singletest},  after the algorithm has gotten stuck in a local minima for some number of iterations where the 
minimum function values are not changing, it is still able to find the global minima at the end. That exemplifies the ability of the algorithm 
to jump out of the local minima and  search for the global minima.
\begin{table}
\centering
\caption{Parameters used for single dimensional test functions}
\label{tsingle}
\begin{tabular}{|c|c|}
\hline
Parameters & Values\\
\hline
number of groups &10\\
\hline
number of population in each group&25\\
\hline
$r_1$(the portion of element)&0.8\\
\hline
$r_2$(the portion of leader)&0.2\\
\hline
$r_3$(the portion of random)&0.0\\
\hline 
\end{tabular}
\end{table}

The first test function used in the optimization  is the Beale Function \cite{355936} shown in Eq.(\ref{beale}).
The global minimum for this function is located at $f(x)=0$, where $ x_1=3$ $x_2=0.5$. 
The search domain was taken to be [-100,100], $-100\le (x_1 ,x_2) \le 100 $. 
Fig.\ref{fig_beale} shows the test result for the Beale Function.

\begin{equation}
\label{beale}
\begin{split}
f_{Beale}(x_{1},x_{2}) = [1.5-x_{1}(1-x_{2})]^2 +
 [2.25-x_{1}(1-x_{2}^2)]^2 
 + [2.625-x_{1}(1-x_{2}^3)]^2.  
\end{split}
\end{equation}

%%%%%%%%%%%%%%%%%%%%%%%%%%%%%%%%

Secondly, the Easom Function\cite{Rachid-genetic} which is shown in Eq.(\ref{easom}) and has a global minimum located at
$f(x)=-1$, where $ x_1=\pi$ $x_2=\pi $. The search domain for this function was taken as  [-100,100], $-100\le (x_1 ,x_2) \le 100 $, in the implementation, and the result is presented in Fig.\ref{fig_easom}.

\begin{equation}
\label{easom}
f_{Easom}(x_{1},x_{2})=-cos(x_{1})cos(x_{2})exp\left(-(x_{1}-\pi)^2-(x_2-\pi)^2\right).
\end{equation}

%%%%%%%%%%%%%%%%%%%%%%%%%%%%%%%%%%
Another single dimensional test function is Goldstein-Price's Function \cite{Rachid-genetic}, Eq.(\ref{gold}).  
The global minimum for this function is at $f(x)=3$ where $ x_1=0$ $x_2=-1$. The search domain used for this problem is [-100,100] $-100\le (x_1 ,x_2) \le 100$, and the result is shown in \ref{fig_gold}.
\begin{equation}
\label{gold}
\begin{split}
f_{GP}(x_1, x_2) = [1+(x_1+x_2+1)^2(19-14x_1+
13x_1^2-14x_2 +6x_1x_2 + 3x_2^2]* \\ [30 + (2x_1-3x_2)^2(18-32x_1 + 12x_1^2+ 48x_2- 36x_1x_2 + 27x_2^2].
\end{split}
\end{equation}

%%%%%%%%%%%%%%%%%%%%%%%%%%%%%%%%%%%
Shubert's Function\cite{Rachid-genetic} is the last single dimensional test function used in the experiments. 
This function, Eq.(\ref{shubert}), has a global minima at $f(x)=-186.7309$ for the search domain [-10,10], $-10\le (x_1 ,x_2) \le 10$.
Fig.\ref{fig_shubert} shows the result for this function.

\begin{equation}
\label{shubert}
\begin{split}
f_{Shubert}(x_1, x_2) = \displaystyle\left (\sum_{i=1}^5 icos[(i+1)x_1+i]\right)
.\left(\sum_{i=1}^5[icos(i+1)x_2+i]\right).
\end{split}
\end{equation}

%%%%%%%%%%%%%%%%%%%%%%%%%%%%%%%%%%%%
%%%%%%%%%%%%%%%%%%%%%%%%%%%%%%%%%%%%%%%%%%%%%%%%%%%%%%%%%%%%%%%%%%%%%%%%%%%%%%%%%

\subsubsection{Multi-Dimensional Test Functions}  
Five multi dimensional test functions were used as test cases. Each of these functions has different properties. 
The number of groups and populations are taken to be the same as in single dimensional functions 
(number of groups is 10 and number of populations in each group is 25). 
The other algorithm parameters and the number of iterations for these functions are given in Table-\ref{tmulti}. 
The number of iterations separately given for each function in the table  were taken to be the same at all dimensions. 
The algorithm was implemented upon each multi dimensional test function at sample dimensions,up to 1000. 
Therefore, for these optimizations, comparisons of running time of GLOA with the number of dimensions are shown in terms of seconds in Fig.\ref{fig_time}
 (Times in Fig.\ref{fig_time} were taken from Matlab which was running on a laptop computer with 1.83GHz Intel Core Duo CPU.).

Multi-dimensional test functions that were used in test cases are as follows:\\

\textit{Rosenbrock's Banana Function \cite{Potter94acooperative,355936,Rachid-genetic,739156}:}
\begin{equation}
f_n(x) = \displaystyle\sum_{i=1}^{n-1}[100(x_i-x_{i+1})^2+(x_i-1)^2].
\end{equation}

The function value at $f_n(x)=0$ is minimum for $x=(1,...,1)$.  The optimization results of the algorithm for this function is shown in
 Fig.\ref{fig_rosen} in which it is seen that the function values even at 1000 dimensions are always less than 1. 
That means the error($\epsilon$) is also less than 1, $\epsilon \le 1$, for all dimensions. Since all multi dimensional 
functions were tested for up to 1000 dimensions; for the following functions we will only give the function descriptions and the error terms.\\
%%%%%%%%%%%%%%%%%%%%%%%%%%%%%%%%%%%%%%

\textit{Griewank's  Function \cite{Potter94acooperative,739156}:}
\begin{equation}
f_n(x) = 1+\displaystyle\sum_{i=1}^{n}\frac{x_i^2}{4000}-\prod_{i=1}^n cos(\frac{x_i}{\sqrt{i}}).
\end{equation}  

 This function is minimum at $f_n(x)=0$, where $x=(0,0,..)$. The results are shown in Fig.\ref{fig_griew}, and $\epsilon \le 0.1$ for all dimensions.
 
%%%%%%%%%%%%%%%%%%%%%%%%%%%%%%%%%%%%%%
\textit{Ackley's  Function \cite{Potter94acooperative}:}
\begin{equation}
\begin{split}
f_n(x) = \displaystyle 20+e-20exp\left ( -0.2\sqrt{\frac{1}{n}\sum_{i=1}^{n}x_i^2}\right )
-exp\left ( \sum_{i=1}^{n}cos(2\pi x_i)\right ). 
\end{split}
\end{equation}

 At $f_n(x)=0$, where $x=(0,0,..)$, the function value is minimum. The results are shown in Fig.\ref{fig_ackley}, and  $\epsilon \le 0.05$ for all dimensions.\\

%%%%%%%%%%%%%%%%%%%%%%%%%%%%%%%%%%%%%%%%
\textit{Sphere Function \cite{739156}:}
\begin{equation}
f_n(x) = \displaystyle \sum_{i=1}^{n}x_i^2. 
\end{equation}

The minimum of the function is located at $f_n(x)=0$, where $x=(0,0,..)$. The results are shown in Fig.\ref{fig_sphere}, and $\epsilon \le 0.1$ for all dimensions.\\ 

\textit{Rastrigin Function \cite{Potter94acooperative, citeulike:3182909}:}
\begin{equation}
f_n(x) = \displaystyle 10n+\sum_{i=1}^{n}\left (x_i^2-10cos\left (2\pi x_i\right ) \right ).
\end{equation}

 At $f_n(x)=0$, where $x=(0,0,..)$, the function value is minimum. The results are shown in Fig.\ref{fig_rastrigin}, and $\epsilon \le 0.05$ for all dimensions.
 
\begin{table*}[!t]
\centering
\caption{Parameters used for multi dimensional test functions}
\label{tmulti}
\begin{tabular}{|c|c|c|c|c|c|}
\hline
Test Function & $r_1$ &$r_2$&$r_3$&domain&number of iteration\\
\hline
Griewank &0.8&0.19&0.01&[-600,600]&1000\\
\hline
Ackley & 0.8&0.19&0.01&[-32.768,32.768]&1000\\
\hline
Rosenbrock &0.8&0.1&0.1&[-100,100]&2000\\
\hline
Sphere &0.6&0.2&0.2 & [-100,100]&1000\\
\hline
Rastrigin&0.6&0.2&0.2 &[-10,10]&1500\\
\hline
\end{tabular}
\end{table*}

\subsection{Lennard-Jones Clusters}

The Lennard-Jones(LJ) potential describes the interaction of a pair of neutral particles. 
The total potential of interactions between N-particles is defined in Eq.(\ref{ljequ}), 
and the pair potential is shown in Fig.\ref{ljpair}. Finding the minimum energy structure of 
 interaction between pairs can be described as an optimization problem. 
The problem is to place the particles in 3-D  space such that their total  interaction 
energy  is at  minimum. 

Global optimization of LJ clusters  is one of the most intensely  studied molecular conformation 
problems since it is simple and accurate enough to describe real  physical interacting  systems. 
Moreover, the simulation results of LJ clusters can be directly compared with the laboratory results [11]. 
Because the number of local minima in the surface of an LJ cluster grows exponentially with N \cite{Stillinger07091984,WalesDoye_JCP119_12409}, 
the solution of the problem is very complex and difficult \cite{Wille}. 
For instance, for the cluster of 13 particles, there are about 1500 local minima \cite{Doye:382358}. 
Thus, finding the minimum energy structure of LJ clusters is still an open and attractive research 
area and a good benchmark problem for the optimization algorithms\cite{739156}. 

The interaction potential is given by:
\begin{equation}
\label{ljequ}
\displaystyle E=4\epsilon \sum_{i<j}^N
 \left[ \left( \frac{\sigma}{r_{ij}} \right)^{12}-\left(\frac{\sigma}{r_{ij}}\right)^{6} \right],
\end{equation}

where $r_{ij}$ is the distance between two particles. For simplicity we will take 
$\epsilon = 1$ and $\sigma = 1$ \cite{wales-1997-101}.

In the implementation, the solution of one  cluster was used as initial guess for a 
larger cluster, this significantly shortened the optimization time (refer to Fig.\ref{fig_ljtime}). 
Here,  the sum of algorithm parameters $r_1$ and $r_2$  is taken  to be  1, 
while $r_3$ is set sufficiently small to locate the values of particles with high precision.  
Table  \ref{table_LJparameters} gives  the values of the  parameters  used in the optimization. 
In addition, as the crossover between groups is done, the transferred elements are mutated. 
This mutation rate is determined also
by the parameter  $r_3$. This mutation does not change the transferred value dramatically; 
however, it helps the algorithm to decrease number of iteration needed for 
small mutations.

The algorithm was run to locate the minimum energy structure  for the clusters of 5-42, 38, 55, 65, and 
75 particles. 
This running choice helped us to observe the growth of the computation time 
with respect to the number of particles. The optimization process was terminated if either the absolute error 
which was defined as the absolute difference between the known lowest (reported in \cite{LJtable}) and the 
found energy level was less than 0.001 or 
the maximum number of iterations (3000) was reached.  The algorithm has been able to find the lowest known 
energy structure
with the absolute errors all of which are less than 0.1. The clusters of 38 and 75 particles are known to  be  
difficult cases  for optimization algorithms. For these cases the first encounter times of the algorithm 
from 100 random starting points (the previous structure is not used 
as an starting point)  are shown 
in Fig.\ref{fig_lj38} and Fig.\ref{fig__lj75} respective to 38 and 75 particles.  The computational  time of the algorithm 
with respect to the 
cluster size is given in Fig.\ref{fig_ljtime}. Thus, 
the scaling of the algorithm with respect to the number of particles is $N^{2.5}$, which is an 
improvement over 
the previous reported scalings: $N^{4.7}$ for the genetic algorithm and $N^{2.9}$ for the pivot
 methods \cite{PhysRevE.55.1162,gregurick:2684}. However, in terms of first encounter time, we see that GLOA in this present format is not as successful as the 
global optimization by basin-hoping for LJ clusters \cite{wales-1997-101}.

\begin{table}[!t]
\caption{Parameters of the Algorithm for the global optimization of LJ clusters}
\label{table_LJparameters}
\centering
\begin{tabular}{|p{2.4in}|p{1in}|}
\hline
Parameters & Values\\
\hline
number of group & 15\\
\hline
number of population in each group & 30\\
\hline
the domain of $r_1$ & $[0.85,0.95]$\\
\hline
the domain of $r_2$ & $[0.15,0.05]$\\
\hline
the domain of $r_3$ &  $[0.001,0.0001]$\\
\hline
initial generated population domain &  [-2,2]\\
\hline
maximum number of iteration & 3000\\
\hline
\end{tabular}
\end{table}
\subsection{Quantum Circuit Design}
In the field of quantum computation, finding quantum circuit designs by decomposing a given unitary matrix
 into a proper-minimum cost quantum gate sequence for the implementation of quantum algorithms and the simulation of molecular systems
 are of fundamental importance.
Evolution of quantum circuits faces two major challenges: complex and huge search space and the high costs of simulating
quantum circuits on classical computers. The optimization task involves not only finding the right sequence of gates,
 but the minimization of the cost of the sequence as well. In the circuit model of quantum computing, each computation or 
algorithm can be defined as a unitary matrix. Thus, the problem becomes the decomposition of a given unitary matrices into a sequence of unitary operators 
which are describing the defined quantum gates.  For this optimization problem, the objective function to be minimized is defined as follows:
\begin{equation}
\label{objective}
y=\left|1-(\alpha C+\frac{\beta}{Cost})\right|,
\end{equation}
 where $C$ is a value to determine the correctness of the circuit; $Cost$ is the implementation cost of the found circuit design;
 and $\alpha$ and $\beta$
are weights to adjust the importance of the correctness and the cost of the circuit in the objective function. The correctness ($C$) is 
defined as $\left|\frac{Tr\left(U_gU_f^\dagger\right)}{N}\right|$, where $U_g$ and $U_f$ are the given and the found unitary matrices of order N, relatively; 
Tr(...) is the trace of a matrix; and N is the $2^n$ (n number of the qubits). When $U_g=U_f$; because all the diagonal elements of the product of $U_g$ and $U_f$ 
becomes ones, the correctness value is one ($C$ is always in the range of one and zero). 
The cost of a circuit is defined as arbitrarily by considering the common implementation costs of  quantum gates in different quantum computer models: 
the number of control qubits and the distance between the target and the control qubits of a gate in the circuits which involve many qubits.
 Hence, the cost of a control gate are determined 
by multiplying the distance (number of qubits) between the target and the control  qubits of the gate by two, and the cost of a single gate is taken one. The cost of a circuit is found by summing 
up the cost value of each quantum gate in the circuit. 
 
In the optimization, $\alpha$ and $\beta$ are considered as 0.9 
and 0.1 which are the best choice among different alternatives to reduce the number of iterations and increase the correctness.  The members in the population of 
the group structured algorithm are taken as genomes which represent a circuit as a numeric string describing gates and their target-control qubits and angles. The order of 
the gates in a genome represents their order in the circuit with respect to the string: The string, \textbf{2} 3 2 0.3 \textbf{3} 2 1 0.5, represents two quantum gates with 
related to the integers \textbf{2} and \textbf{3} with their target and control qubits and their angle values: 3 2 0.3 and 2 1 0.5, respectively.
%%%%%%%%%%%%%%%%%%%%%%%%%%%%%%%%%%%%%%%%%%%%%%%%%%%

As a test case, we use the Grover search algorithm \cite{Grover} which is one of the advances quantum computing 
has brought on classical computing. 
It reduces the computational time of a brute force search from $O(N)$ to $O(\sqrt{N})$. 
The algorithm can be described in four steps \cite{Nielsen,Jing-Paper}:
 \begin{enumerate}
  \item Start with an n-qubit initial state $|000..0\rangle$.
\item Put this initial state into the superposition by applying Hadamard (H) gates to the each qubit.
\item $\lfloor\frac{\pi}{4}\sqrt{N}\rfloor$ times:
\begin{itemize}
\item Apply the first operator $ U_f$ which is defined as:
    \begin{equation}
  U_f=I-|a\rangle\langle a|,  
    \end{equation}
where $a$ is the element that is being searched, and $I$ is the identity.
The act of $U_f$ is to mark the element $x$ if and only if $x=a$; the function, $f(x)$, is equal to 1 for $x=a$.
\item Apply the second operator (the inversion about the average operator or diffusion operator) $U_{\psi^{\perp}}$ which is defined as:
\end{itemize} 
\begin{equation}
 U_{\psi^{\perp}}=2|\psi\rangle\langle\psi|-I.
\end{equation}
$U_{\psi^{\perp}}$ amplifies the amplitude of the state marked by the first operator
and reduces the amplitudes of the rest. Thus, the probability of seeing the marked 
element at the end of measurement gets higher. The matrix D representing this operator is found as follows:
\begin{equation}
\label{groveroperator}
    D_{ij}=\left\{\begin{array}{cc}
                                        \frac{N}{2},& if\ i\ne j\\
                                        -1 + \frac{N}{2},& if \ i=j
                                        \end{array}\right\}.
\end{equation}
\item Measure the result.
 \end{enumerate}

\begin{figure}[!t]
\centering
\includegraphics[width=0.6\textwidth]{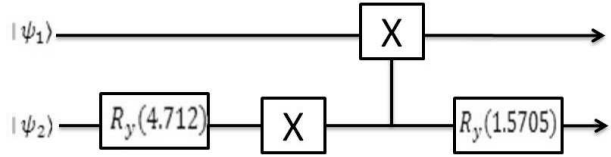}
\caption{Quantum circuit design for the inversion about the average part of the Grover search algorithm.}
\label{fig_grover}
\end{figure}

\begin{figure*}[!t]
\centerline{
\subfigure[Rosenbrock Function]{\includegraphics[width=2.5in]{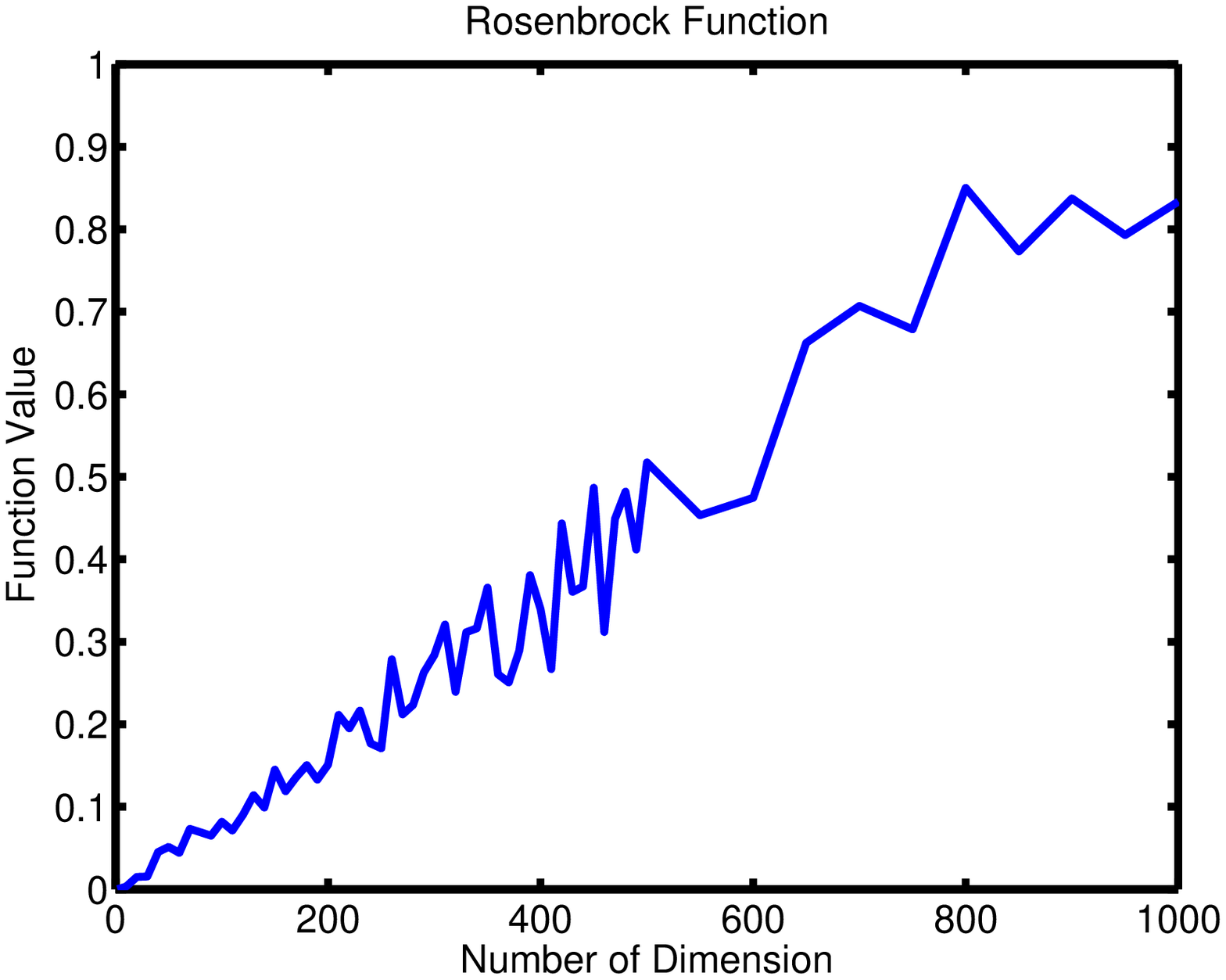}
\label{fig_rosen}}
\hfil
\subfigure[Griewank Function]{\includegraphics[width=2.5in]{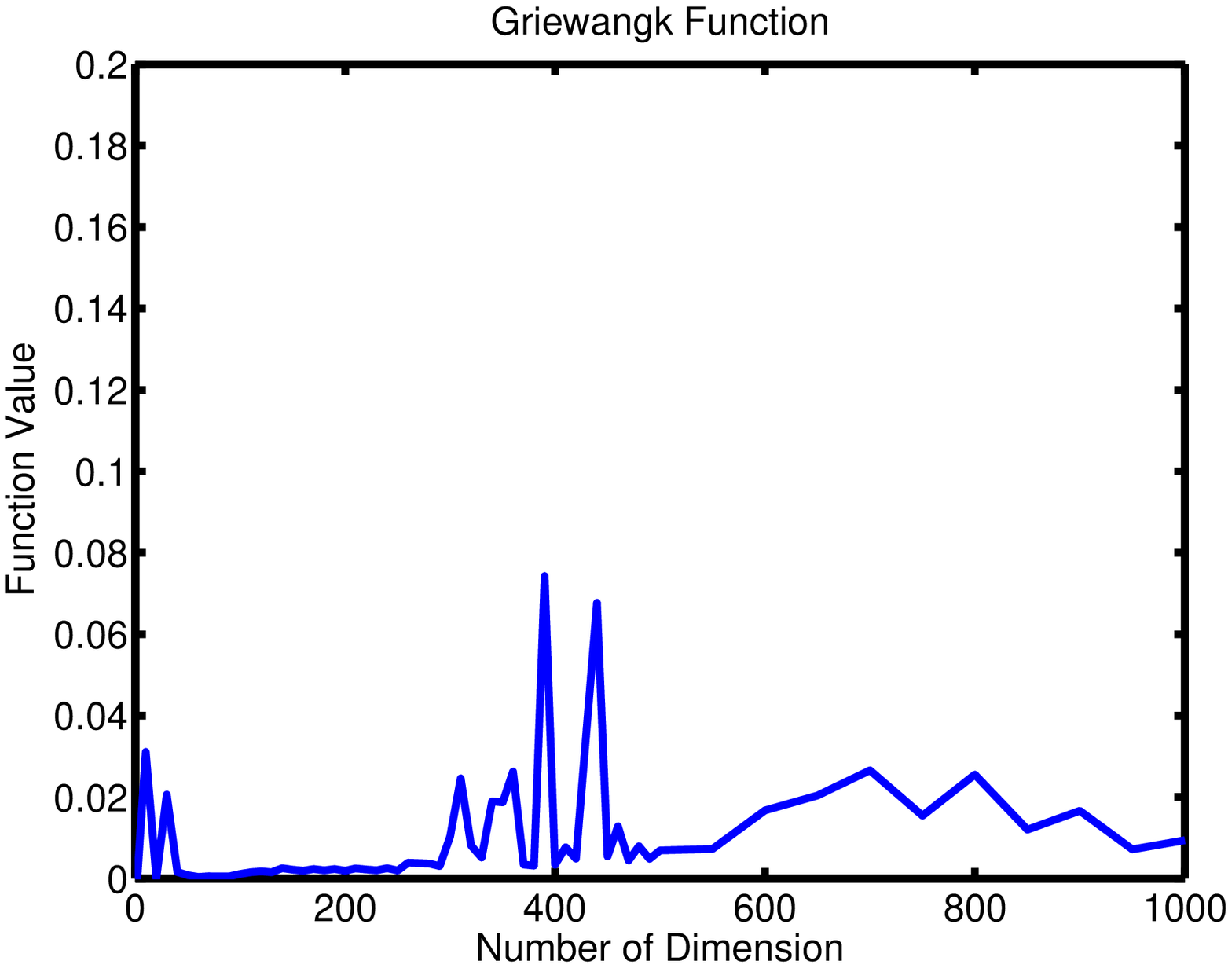}
\label{fig_griew}}}
\centerline{
\subfigure[Ackley Function]{\includegraphics[width=2.5in]{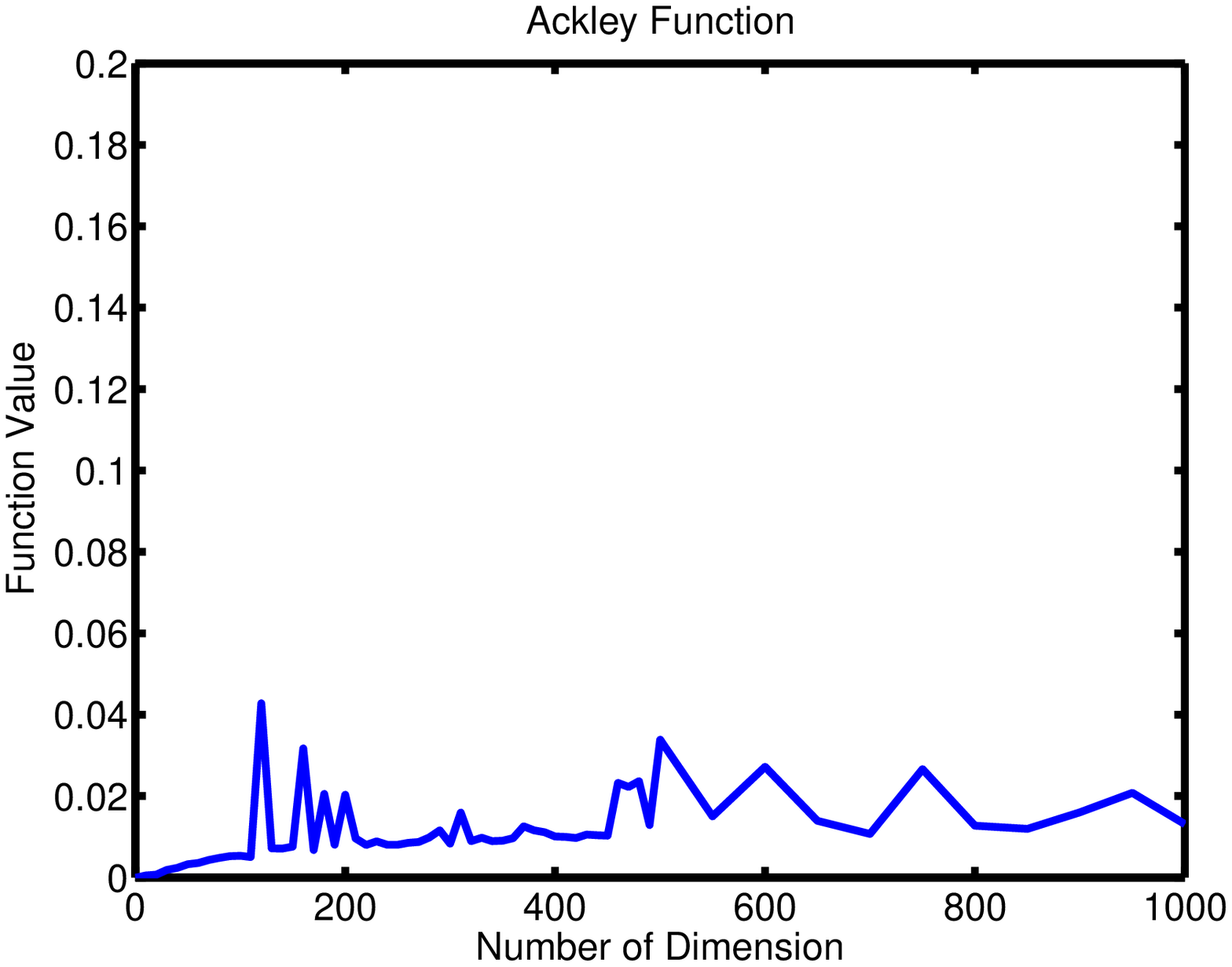}
\label{fig_ackley}}
\hfil
\subfigure[Sphere Function]{\includegraphics[width=2.5in]{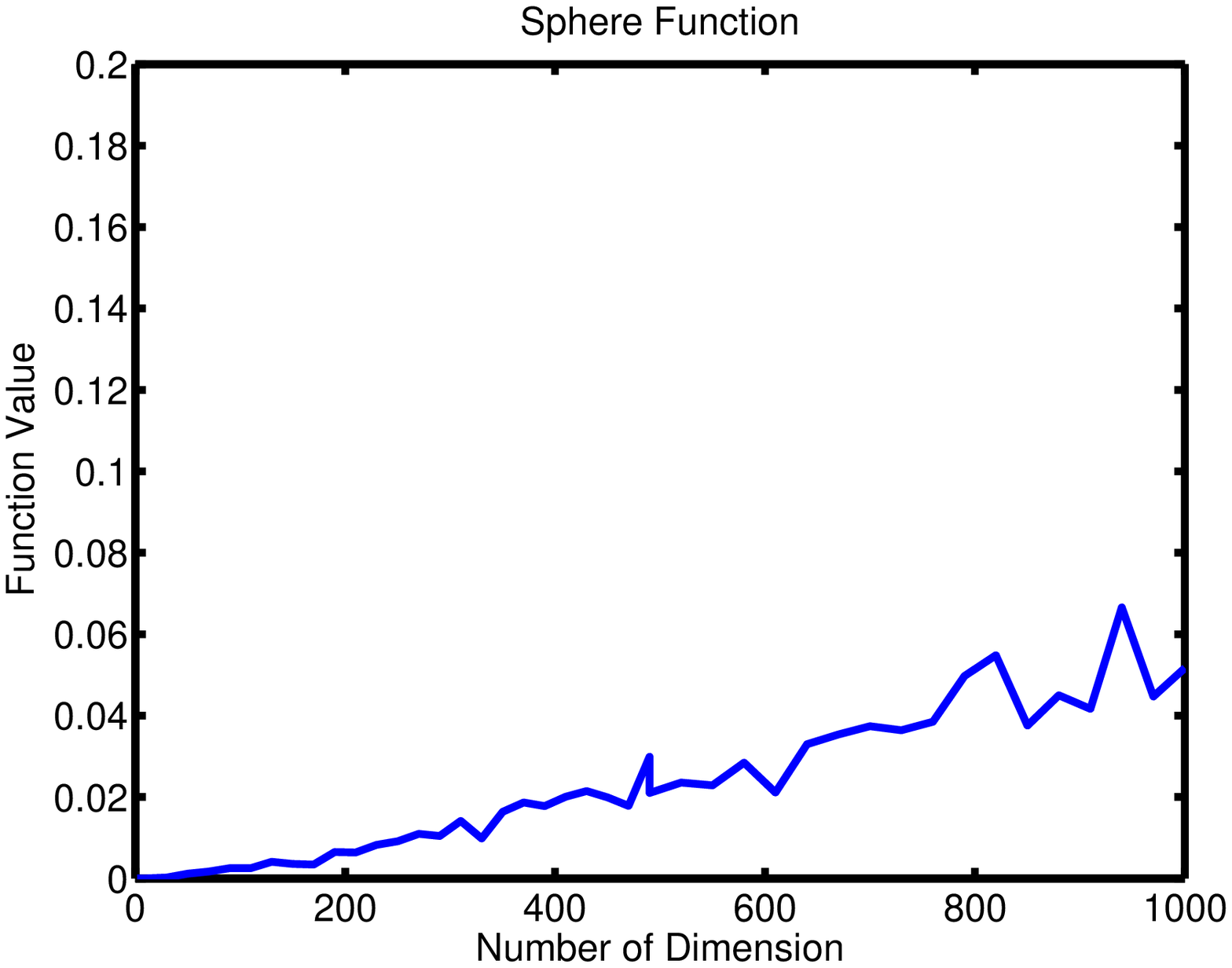}
\label{fig_sphere}}}
\centerline{
\subfigure[Rastrigin Function]{\includegraphics[width=2.5in]{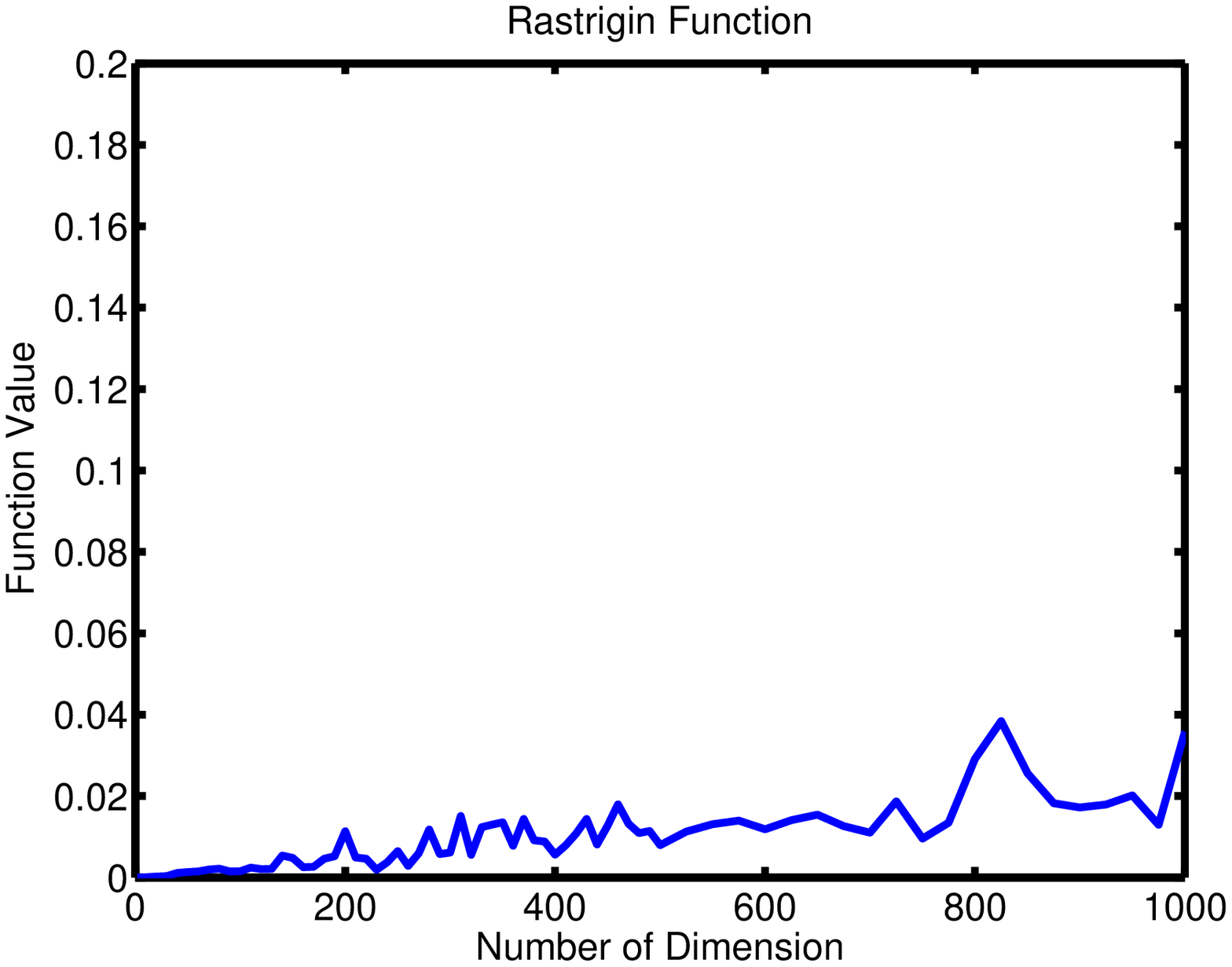}
\label{fig_rastrigin}}
\hfil
\subfigure[Running Time]{\includegraphics[width=2.5in]{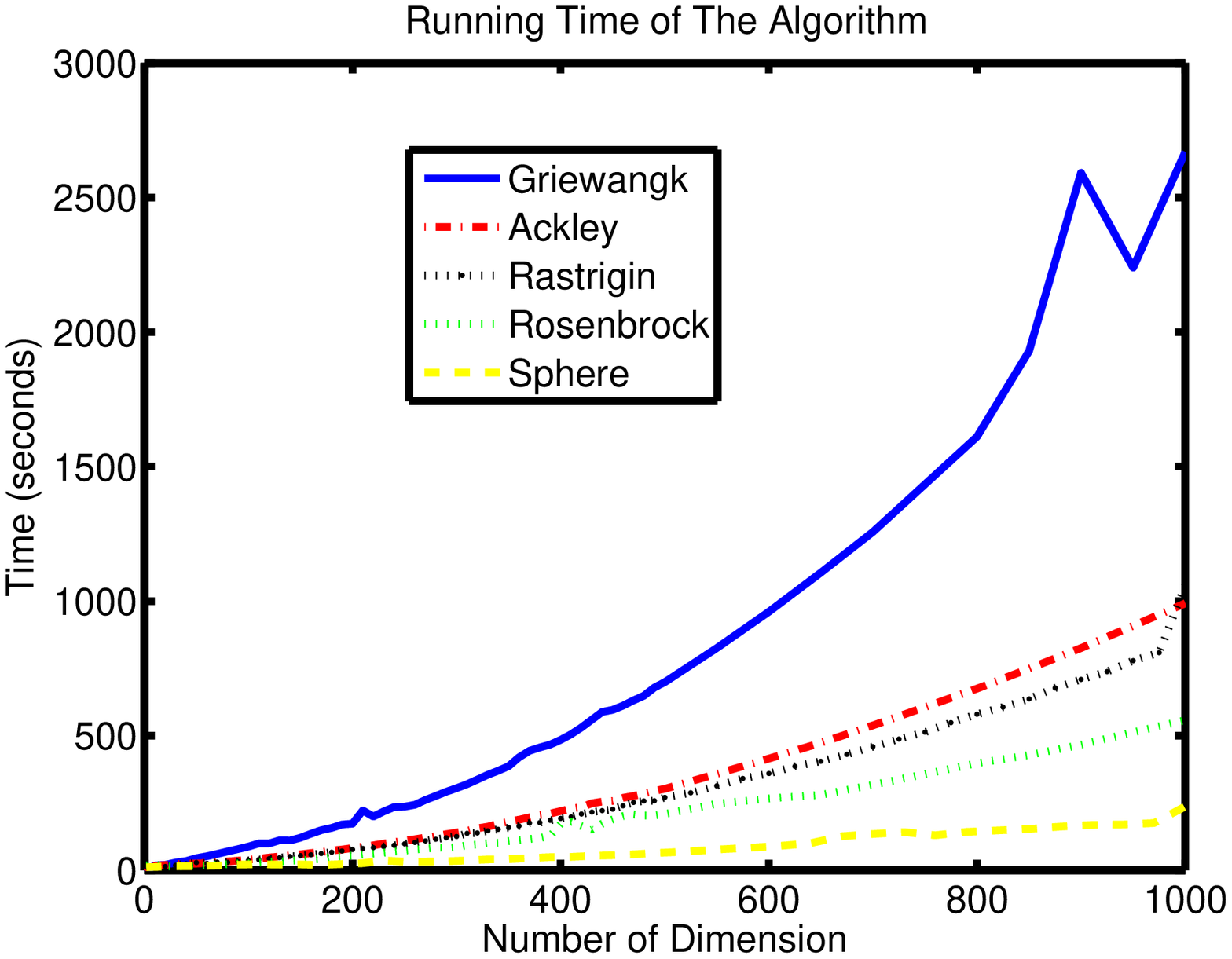}
\label{fig_time}}}
\caption{Results For Multi Dimensional Optimization Test Functions: 
for (a)-(e), the x-axis is the number of dimension, 
and the y-axis is the function value which is the result of the best member in the population. 
 For (f), the x-axis is the number of dimension, and the y-axis is the running time of the algorithm. 
(a) shows the result for Rosenbrock Function; (b) is the result for Griewank Function; (c) is the result for Ackley Function; 
(d) is the result for Sphere Function; (e) is the result for Rastrigin Function; and (f) shows the CPU running time of the algorithm with 
respect to the dimension of the problems.}
\label{multitest}
\end{figure*}

\begin{figure*}[!t]
\centerline{
\subfigure[Lennard-Jones pair potential]{\includegraphics[width=2.65in]{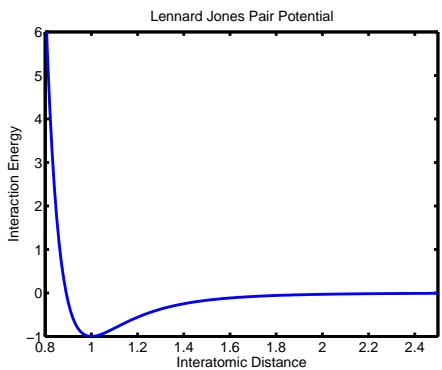}
\label{ljpair}}
\hfil
\subfigure[CPU running times with respect to the clusters]{\includegraphics[width=2.75in]{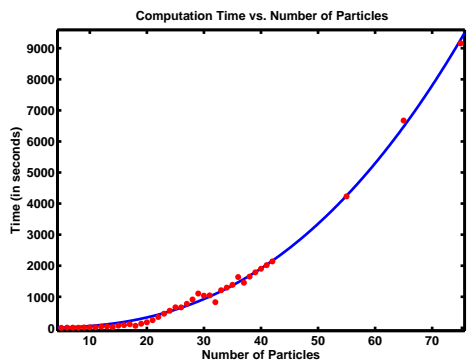}
\label{fig_ljtime}}}
\centerline{
\subfigure[Optimization of the cluster of 38 particles]{\includegraphics[width=2.75in]{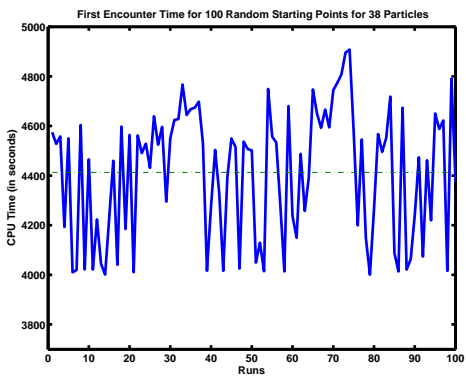}
\label{fig_lj38}}
\hfil
\subfigure[Optimization of the cluster of 75 particles]{\includegraphics[width=2.75in]{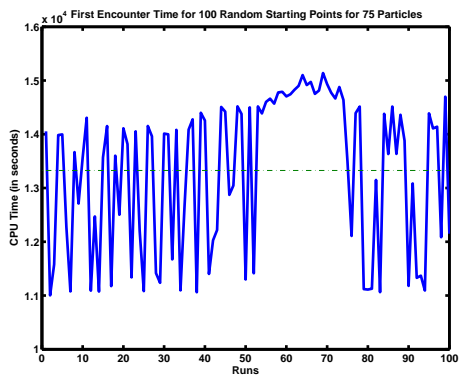}
\label{fig__lj75}}}
\caption{In (a), Lennard-Jones pair potential are shown for two particles which is the minimum at atomic distance 1. In (b), the dots are the measured CPU time for the
 cluster of different number of particles; the solid line which grows with the order of $O(n^{2.5})$ is the fitted curve for the measured data (n is the number of particles).
(c) and (d) show the first encounter time of the algorithm for 100 random starting points for the clusters of 38 and 75 particles.}
\label{fig_lj}
\end{figure*}

%%%%%%%%%%%%%%%%%%%%%%%%%%%%%%%%%%%%%%%%%%%%%%%%%%%%%%%%%%%%%%%%%%%%%%%%%%%%%%%%
 The exact circuit design in Fig.\ref{fig_grover} for the second part (inversion about the average) of the Grover 
search algorithm (matrix elements of which are -0.5 in diagonal and 0.5 in the rest) is found with the objective function value 0.08 by applying the
 algorithm with the parameters given in Table \ref{table_qc}. 

\begin{table}[!t]
\caption{Parameters of the Algorithm for Finding Quantum Circuits}
\label{table_qc}
\centering
\begin{tabular}{|p{1.7in}|p{0.8in}|}
\hline
Parameters& Values\\
\hline
number of group & 15\\
\hline
population in each group & 25\\
\hline
$r_1$ & 0.8\\
\hline
$r_2$ & 0.1\\
\hline
$r_3$ &  0.1\\
\hline
number of iteration & 1000\\
\hline
\end{tabular}
\end{table}

\section{Conclusions}

In this paper we have presented the Group Leaders Optimization Algorithm (GLOA). 
The algorithm  is  quite efficient, very
flexible, rarely gets trapped in local minima, does not require
computationally expensive derivatives, and is quite easy to
implement. Flexibility is further enhanced by the ability to
incorporate any previous knowledge of the potential under
investigation into the optimization. As a specific example,
we used smaller Lennard-Jones clusters as the starting point
for the larger ones. Reducing the computational cost to scale approximately 
as $N^{2.5}$ is very promising to treat larger complex systems.

Recently, in a  promising new direction, we demonstrate that a modified Grovers quantum
algorithm can be applied to real problems of finding a global minimum using modest 
numbers of quantum bits\cite{Jing-Paper}.
Calculations of the global minimum of simple test functions and Lennard-Jones 
clusters have been carried out on
a quantum computer simulator using a modified Grovers algorithm. 
The number of function evaluations N reduced from O(N) in classical 
simulation to O(N$^{1/2}$) in quantum simulation. We also show how the Grovers
quantum algorithm can be combined with the classical Pivot method for 
global optimization to treat larger
systems. We currently investigating combining this new algorithm, 
the Group Leaders Optimization Algorithm, with the Grover's quantum algorithm 
for global optimization of complex systems. The algorithm defined here also allows us to design quantum circuits for further quantum algorithms 
and the simulations of molecular Hamiltonians such as the Hamiltonians of $H_2O$ and $H_2$.  

Therefore, because of the parallel group structure of the algorithm, it can be easily adapted
to a parallel processing environment to improve the efficiency of the algorithm for hard problems. In that case,  each of the groups 
can be run on a different machine and the parameter transfer between groups can be done through either a shared memory 
or a message-passing interface. 

\section{Acknowledgments}
We would like  to thank the Army Research Office (ARO) 
and the NSF Center for Quantum Information and Computation for 
Chemistry, award number CHE-1037992, for financial support of this project.

\bibliographystyle{tMPH}   
\bibliography{./paper}

\begin{thebibliography}{40}
\providecommand{\url}[1]{\texttt{#1}}
\providecommand{\urlprefix}{URL }
\markboth{Taylor \& Francis and I.T. Consultant}{Molecular Physics}

\bibitem{W2009GOEB}
T. Weise, \emph{Global Optimization Algorithms -- Theory and Application}
  (Thomas Weise, University of Kassel, Germany, 2007).
  $<${http://www.it-weise.de/projects/book.pdf}$>$.

\bibitem{Wille}
L.T. Wille and J. Vennik, Computational complexity of the ground-state
  determination of atomic clusters\ Journal of Physics A: Mathematical and
  General \textbf{18} (8), L419 (1985).

\bibitem{Adib}
A.B. Adib, NP-hardness of the cluster minimization problem revisited\ Journal
  of Physics A: Mathematical and General \textbf{38} (40), 8487 (2005).

\bibitem{Dennis}
J.E. Dennis and R.B. Schnabel, \emph{{Numerical Methods for Unconstrained
  Optimization and Nonlinear Equations (Classics in Applied Mathematics)}}
  (Society for Industrial Mathematics, Philadelphia, PA, USA, 1996).

\bibitem{Gilbert}
J.C. Gilbert and C. Lemar{\'e}chal, Some numerical experiments with
  variable-storage quasi-Newton algorithms\ Math. Program. \textbf{45} (3),
  407--435 (1989).

\bibitem{Levy}
A.V. Levy and A. Montalvo, The Tunneling Algorithm for the Global Minimization
  of Functions\ SIAM Journal on Scientific and Statistical Computing \textbf{6}
  (1), 15--29 (1985).

\bibitem{Shalloway}
D. Shalloway, Application of the renormalization group to deterministic global
  minimization of molecular conformation energy functions\ Journal of Global
  Optimization \textbf{2}, 281--311 (1992).

\bibitem{Kirkpatrick}
S. Kirkpatrick, Optimization by simulated annealing: Quantitative studies\
  Journal of Statistical Physics \textbf{34}, 975--986 (1984).

\bibitem{Finnila}
A.B. Finnila, M.A. Gomez, C. Sebenik, C. Stenson and J.D. Doll, Quantum
  annealing: A new method for minimizing multidimensional functions\ Chemical
  Physics Letters \textbf{219} (5-6), 343--348 (1994).

\bibitem{Frantz}
D.D. Frantz, D.L. Freeman and J.D. Doll, Reducing quasi-ergodic behavior in
  Monte Carlo simulations by J-walking: Applications to atomic clusters\ The
  Journal of Chemical Physics \textbf{93} (4), 2769--2784 (1990).

\bibitem{Cvijovicacute}
D. Cvijovicacute and J. Klinowski, {Taboo Search: An Approach to the Multiple
  Minima Problem}\ Science \textbf{267} (5198), 664--666 (1995).

\bibitem{Goldberg}
D.E. Goldberg, \emph{Genetic Algorithms in Search, Optimization and Machine
  Learning}   (Addison-Wesley Longman Publishing Co., Inc., Boston, MA, USA,
  1989).

\bibitem{Wales27081999}
D.J. Wales and H.A. Scheraga, {Global Optimization of Clusters, Crystals, and
  Biomolecules}\ Science \textbf{285} (5432), 1368--1372 (1999).

\bibitem{wales-1997-101}
D. Wales and J. Doye, Global Optimization by Basin-Hopping and the Lowest
  Energy Structures of Lennard-Jones Clusters Containing up to 110 Atoms\
  J.PHYS.CHEM.A \textbf{101}, 5111 (1997).

\bibitem{229867}
T. B{\"a}ck, \emph{Evolutionary algorithms in theory and practice: evolution
  strategies, evolutionary programming, genetic algorithms}   (Oxford
  University Press, Oxford, UK, 1996).

\bibitem{PhysRevE.55.1162}
P. Serra, A.F. Stanton and S. Kais, Pivot method for global optimization\ Phys.
  Rev. E \textbf{55} (1), 1162--1165 (1997).

\bibitem{Kennedy95}
J. Kennedy and R. Eberhart, Particle swarm optimization. in \emph{Neural
  Networks, 1995. Proceedings., IEEE International Conference on}, Vol.~4,
  August.  $<${http://dx.doi.org/10.1109/ICNN.1995.488968}$>$, pp. 1942--1948.

\bibitem{StantonBleilKais}
A.F. Stanton, R.E. Bleil and S. Kais, A new approach to global minimization\
  Journal of Computational Chemistry \textbf{18} (4), 594--599 (1997).

\bibitem{SerraStantonKais}
P. Serra, A. Stanton, S. Kais and R. Bleil, Comparison study of pivot methods
  for global optimization\ Journal of Chemical Physics \textbf{106} (17),
  7170--7177 (1997).

\bibitem{Nigra1999433}
P. Nigra and S. Kais, Pivot method for global optimization: a study of water
  clusters \uppercase{$(H\_2O)\_N$} \uppercase{N} with 2 $\leq$ \uppercase{N}
  $\leq$ 33\ Chemical Physics Letters \textbf{305} (5-6), 433--438 (1999).

\bibitem{Weinberg}
B. Weinberg, The Effect of Group Leaders, in \emph{2008-09 Mershon Center
  Research Projects (Institutions that Manage Violent Conflict)}, OH, USA.
  $<${http://hdl.handle.net/1811/36219}$>$.

\bibitem{Potter94acooperative}
M.A. Potter and K.A.D. Jong, A Cooperative Coevolutionary Approach to Function
  Optimization. in   (Springer-Verlag, London, UK, 1994), pp. 249--257.

\bibitem{citeulike:3182909}
F. {van den Bergh} and A.P. Engelbrecht, A Cooperative approach to particle
  swarm optimization\ Evolutionary Computation, IEEE Transactions on \textbf{8}
  (3), 225--239 (2004).

\bibitem{517269}
E. Cantu-Paz, \emph{Efficient and Accurate Parallel Genetic Algorithms}
  (Kluwer Academic Publishers, Norwell, MA, USA, 2000).

\bibitem{1108890}
M.A. Potter and K.A. {De Jong}, Cooperative Coevolution: An Architecture for
  Evolving Coadapted Subcomponents\ Evol. Comput. \textbf{8} (1), 1--29 (2000).

\bibitem{Yang20082985}
Z. Yang, K. Tang and X. Yao, Large scale evolutionary optimization using
  cooperative coevolution\ Information Sciences \textbf{178} (15), 2985--2999
  (2008), Nature Inspired Problem-Solving.

\bibitem{WinGeiGal}
G. Winter, D. Geiner and B. Galvan, Parallel evolutionary computation, Spain
  2005.

\bibitem{Turner2000183}
G.W. Turner, E. Tedesco, K.D.M. Harris, R.L. Johnston and B.M. Kariuki,
  Implementation of Lamarckian concepts in a Genetic Algorithm for structure
  solution from powder diffraction data\ Chemical Physics Letters \textbf{321}
  (3-4), 183--190 (2000).

\bibitem{DifEvolution}
R. Storn and K. Price, Differential Evolution - A Simple and Efficient
  Heuristic for global Optimization over Continuous Spaces\ Journal of Global
  Optimization \textbf{11}, 341--359 (1997).

\bibitem{355936}
J.J. Mor{\'e}, B.S. Garbow and K.E. Hillstrom, Testing Unconstrained
  Optimization Software\ ACM Trans. Math. Softw. \textbf{7} (1), 17--41 (1981).

\bibitem{Rachid-genetic}
R. Chelouah and P. Siarry, Genetic and Nelder-Mead algorithms hybridized for a
  more accurate global optimization of continuous multiminima functions\
  European Journal of Operational Research \textbf{148} (2), 335--348 (2003).

\bibitem{739156}
P.J. Angeline, Evolutionary Optimization Versus Particle Swarm Optimization:
  Philosophy and Performance Differences. in \emph{EP '98: Proceedings of the
  7th International Conference on Evolutionary Programming VII}
  (Springer-Verlag, London, UK, 1998), pp. 601--610.

\bibitem{Stillinger07091984}
F.H. Stillinger and T.A. Weber, {Packing Structures and Transitions in Liquids
  and Solids}\ Science \textbf{225} (4666), 983--989 (1984).

\bibitem{WalesDoye_JCP119_12409}
D.J. Wales and J.P.K. Doye, {Stationary points and dynamics in high-dimensional
  systems}\ J. Chem. Phys. \textbf{119}, 12409--12416 (2003).

\bibitem{Doye:382358}
J.P.K. Doye, M. Miller and D. Wales, Evolution of the Potential Energy Surface
  with Size for Lennard-Jones Clusters\ J. Chem. Phys. \textbf{111}, 8417--8428
  (1999).

\bibitem{LJtable}
D.J. Wales, J.P.K. Doye, A. Dullweber, M.P. Hodges, F.Y. Naumkin, F. Calvo, J.
  Hern{\'a}ndez-Rojas and T.F. Middleton, The Cambridge Cluster Database .
  $<${http://www-wales.ch.cam.ac.uk/CCD.html}$>$.

\bibitem{gregurick:2684}
S.K. Gregurick, M.H. Alexander and B. Hartke, Global geometry optimization of
  (Ar)[sub n] and B(Ar)[sub n] clusters using a modified genetic algorithm\ The
  Journal of Chemical Physics \textbf{104} (7), 2684--2691 (1996).

\bibitem{Grover}
L.K. Grover, A fast quantum mechanical algorithm for database search. in
  \emph{STOC '96: Proceedings of the twenty-eighth annual ACM symposium on
  Theory of computing}, Philadelphia, Pennsylvania, United States  (ACM, New
  York, NY, USA, 1996), pp. 212--219.

\bibitem{Nielsen}
M.A. Nielsen and I.L. Chuang, \emph{Quantum Computation and Quantum
  Information}, 1st ed.   (Cambridge University Press, Cambridge, 2000).

\bibitem{Jing-Paper}
J. Zhu, Z. Huang and S. Kais, Simulated quantum computation of global minima\
  Molecular Physics: An International Journal at the Interface Between
  Chemistry and Physics \textbf{107}, 2015--2023 (2009).

\end{thebibliography}
%%%%%%%%%%%%%%%%%%%%%%%%%%%%%%%%%
\end{document}